\pdfoutput=1

\documentclass[11pt]{article}

\usepackage{acl}

\usepackage{times}
\usepackage{latexsym}

\usepackage[T1]{fontenc}

\usepackage[utf8]{inputenc}

\usepackage{microtype}

\usepackage{inconsolata}

\usepackage{graphicx}
\usepackage{booktabs} 
\usepackage{amssymb} 
\usepackage{multirow} 
\usepackage{bbding}

\usepackage[most]{tcolorbox}
\usepackage{xcolor}
\usepackage{lipsum}

\definecolor{reflcolor}{HTML}{E8F0FE}

\definecolor{assistantcolor}{HTML}{E6F4EA}

\definecolor{summcolor}{HTML}{E89999}

\newtcolorbox{assistantbox}{
  colback=assistantcolor,
  colframe=green!40!black,
  title=\textbf{System},
  sharp corners,
  boxrule=0.5pt,
}

\newtcolorbox{summarizationbox}{
  colback=summcolor,
  colframe=red!40!black,
  title=\textbf{Summarize},
  sharp corners,
  boxrule=0.5pt,
}

\newtcolorbox{reflectionbox}{
  colback=reflcolor,
  colframe=blue!40!black,
  title=\textbf{Reflect},
  sharp corners,
  boxrule=0.5pt,
}

\title{Increasing the Thinking Budget is Not All You Need}

\author{Ignacio Iacobacci\thanks{Corresponding author \texttt{\href{iiacobacci@elm.sa}{iiacobacci@elm.sa}}.}\thanks{General inquiries \texttt{\href{research@elm.sa}{research@elm.sa}}.}, Zhaozhi Qian, Faroq AL-Tam, \\ \textbf{Muhammad AL-Qurishi}, and \textbf{Riad Souissi} \\
  Elm Company, Digital City, Riyadh 12382, KSA}

\begin{document}
\maketitle
\begin{abstract}

Recently, a new wave of thinking-capable Large Language Models has emerged, demonstrating exceptional capabilities across a wide range of reasoning benchmarks. Early studies have begun to explore how the amount of compute in terms of the length of the reasoning process, the so-called \textit{thinking budget}, impacts model performance. In this work, we propose a systematic investigation of the thinking budget as a key parameter, examining its interaction with various configurations such as self-consistency, reflection, and others. Our goal is to provide an informative, balanced comparison framework that considers both performance outcomes and computational cost.
Among our findings, we discovered that simply increasing the thinking budget is not the most effective use of compute. More accurate responses can instead be achieved through alternative configurations, such as self-consistency and self-reflection.

\end{abstract}

\section{Introduction}

Large language models (LLMs) have made remarkable strides in recent years, achieving unprecedented levels of performance in complex reasoning tasks.
Recent efforts have been designed not only to produce answers, but also to show an explicit step-by-step reasoning process that leads to the answer \cite{openAI-o1-preview,anthropic-claude,deepmind-gemini,deepseekai2025deepseekr1incentivizingreasoningcapability,qwen3,meta-llama-4} .

Some works have been exploring how these models behave when their \textit{thinking budget}--the maximum amount of thinking tokens-- is taken into account and its relation with model's performance. \citet{muennighoff2025s1simpletesttimescaling} for instance investigated the scaling, showing that increasing compute during inference improves overall performance of the models. However, little is known about how these reasoning models perform in more \textit{agentic} configurations.
On the other hand, \citet{wang-etal-2024-reasoning-token} perform a thoughtful analysis comparing different reasoning approaches such as single agent, multi-agent, reflection, self-consistency, etc, from the perspective of the reasoning budget, but without dealing with explicit reasoning models.

Nevertheless, there has been no effort yet to understand how the amount of compute performs depending on the configuration where it is used. In this work, we design a complete suite of configurations that, together with the compute budget, provide a balanced perspective on reasoning strategy effectiveness. These configurations vary across multiple dimensions, including model size, inference-time optimization techniques, allowing us to isolate and examine the role that compute plays in different operational settings. By systematically evaluating these variables, we aim to uncover nuanced interactions between compute allocation and reasoning performance, ultimately guiding more efficient and adaptive deployment strategies for intelligent systems.

We compare several established reasoning strategies, such as Self-Consistency and Reflection, and explore how these methods perform when combined with varying levels of thinking budgets on challenging benchmarks. By systematically varying the amount of compute allocated to each strategy, we aim to understand how resource distribution influences reasoning quality and decision-making accuracy. Our empirical results unveil compelling insights into how and where to assign computational budgets to maximize overall performance, highlighting trade-offs and synergies across different strategies and tasks.

\noindent Our contributions are threefold:
\begin{itemize}
    \item We introduce a comprehensive framework that evaluates reasoning strategies under diverse compute configurations, providing a structured view of performance trade-offs.
    \item We present extensive empirical results on a set of challenging benchmarks, identifying optimal budget allocation patterns across reasoning techniques. We will show that straightforward extending the thinking budget might not be the best strategy.
    \item We offer actionable guidelines for effectively combining compute and reasoning strategies to achieve better outcomes in real-world AI applications.
\end{itemize}

\section{Past work}

Since the introduction of GPT-3 \cite{brown2020language}, many efforts have been made to improve the capabilities of the models by applying different prompting techniques. \citet{chainofthought} introduced \textit{Chain of thought} (CoT), a technique that later became the de facto initiation technique of LLMs. Empirical results show that CoT improves reasoning capability by encouraging models to plan step-by-step and reason toward the answer, using computation to decompose tasks into smaller and simpler steps. Later, \citet{taylor2022galacticalargelanguagemodel} introduced Galactica, an LLM specifically trained to be used in the science domain. Their training data included not only scientific text, but also LaTeX code from scientific publications, programming code, DNA sequences, etc. Galactica also introduced the idea of having an explicit portion of the generation for step-by-step reasoning, called working memory context. This generation portion was surrounded by a special token \textit{<work>}. 
This idea was further exploited by \citet{openAI-o1-preview} with the introduction of o1-preview, a model designed to \textit{spend more time thinking} before responding. However, this model and the successors do not make the thinking process public. Recently, \citet{deepseekai2025deepseekr1incentivizingreasoningcapability} released the DeepSeek-R1 family of models, an open-source family trained specifically to reward answers that format its thinking process between \textit{<think>} and \textit{</think>} tags.

\begin{figure}[t!]
	\centering
	\includegraphics[width=1.0\linewidth]{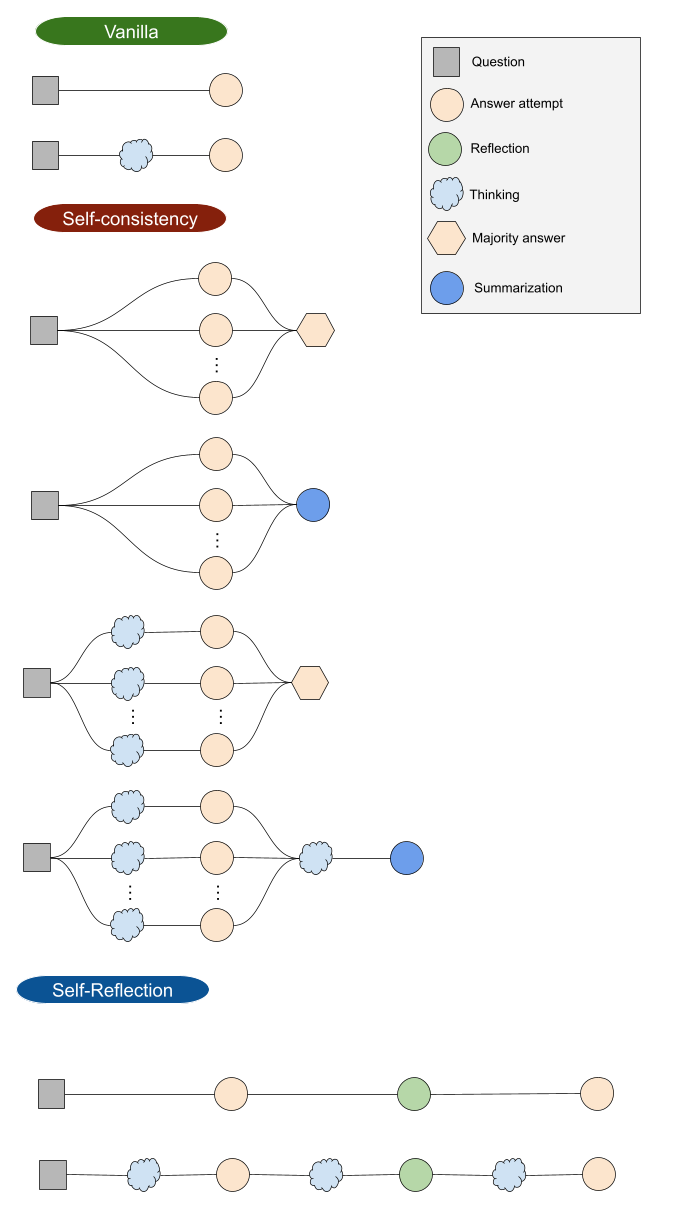}
	\caption{Visualization of reasoning strategies across different configurations.}
	\label{fig:reasoning_models}
\end{figure}

\begin{figure*}[t!]
	\centering
  \begin{minipage}[b]{0.30\textwidth}
    \includegraphics[width=\textwidth]{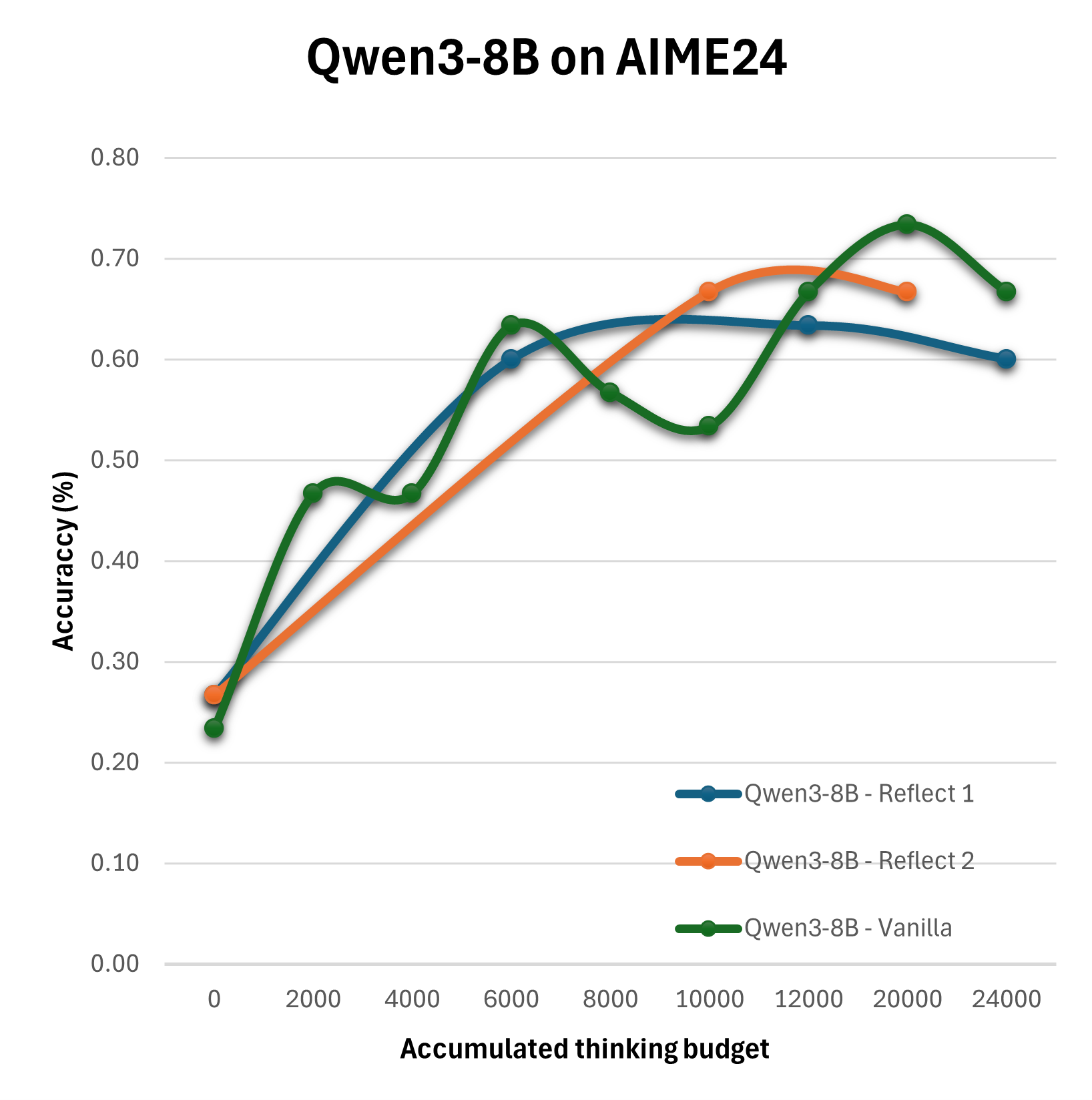}
  \end{minipage}
  \hfill
  \begin{minipage}[b]{0.30\textwidth}
    \includegraphics[width=\textwidth]{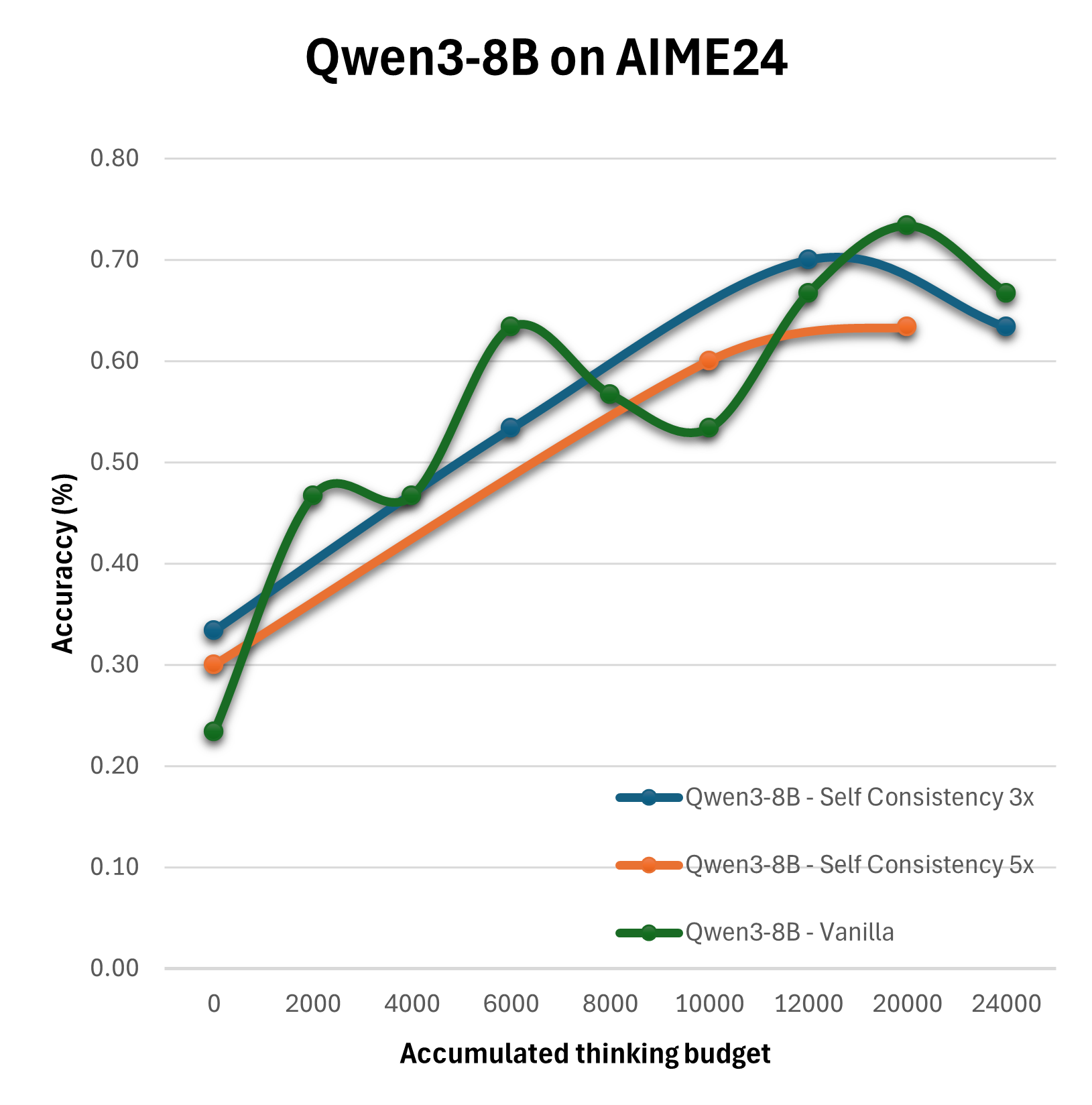}
  \end{minipage}
    \hfill
   \begin{minipage}[t]{0.30\textwidth}
    \includegraphics[width=\textwidth]{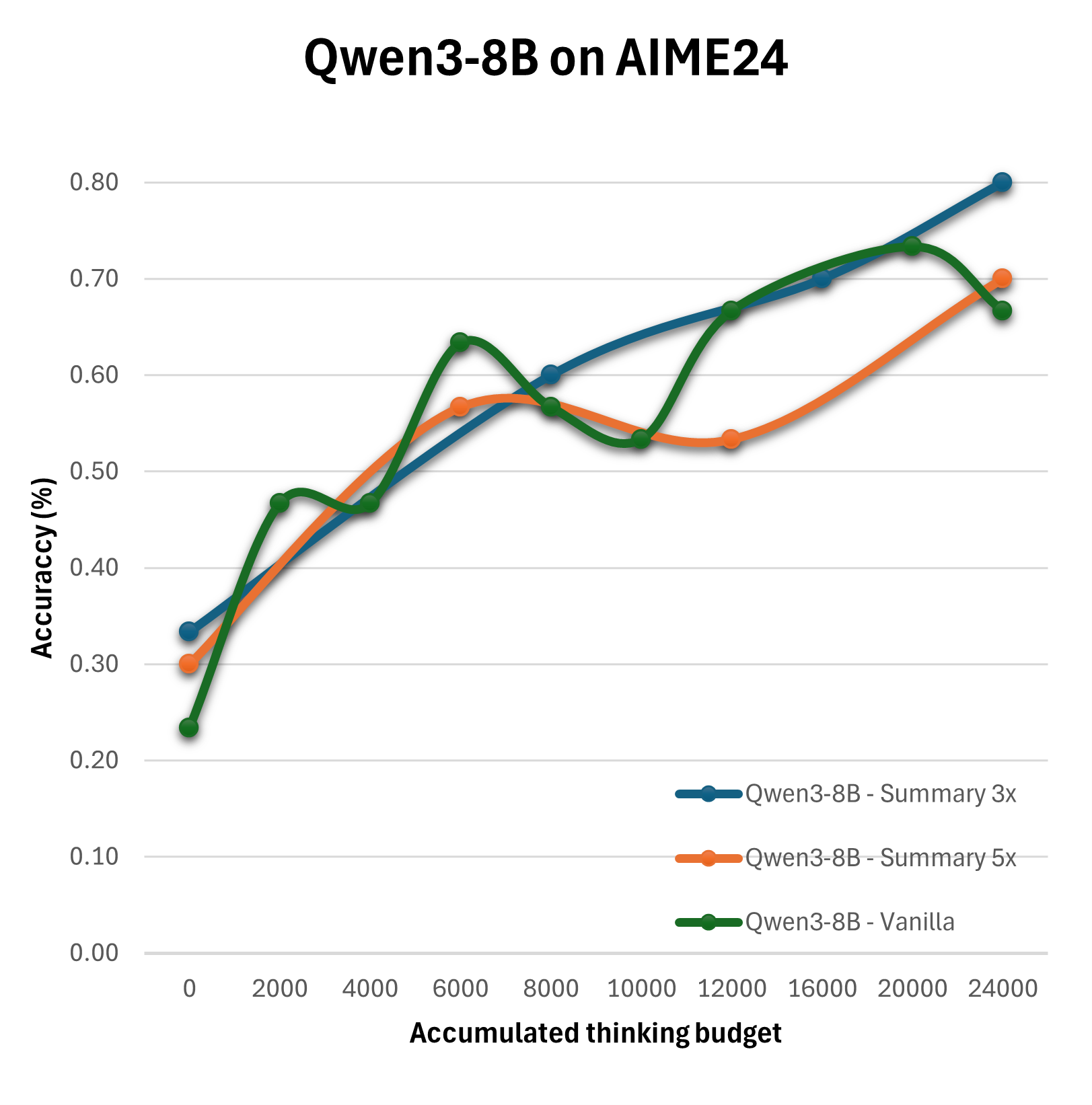}
  \end{minipage}
  \vspace{-0.2cm}
  \begin{minipage}[b]{0.30\textwidth}
    \includegraphics[width=\textwidth]{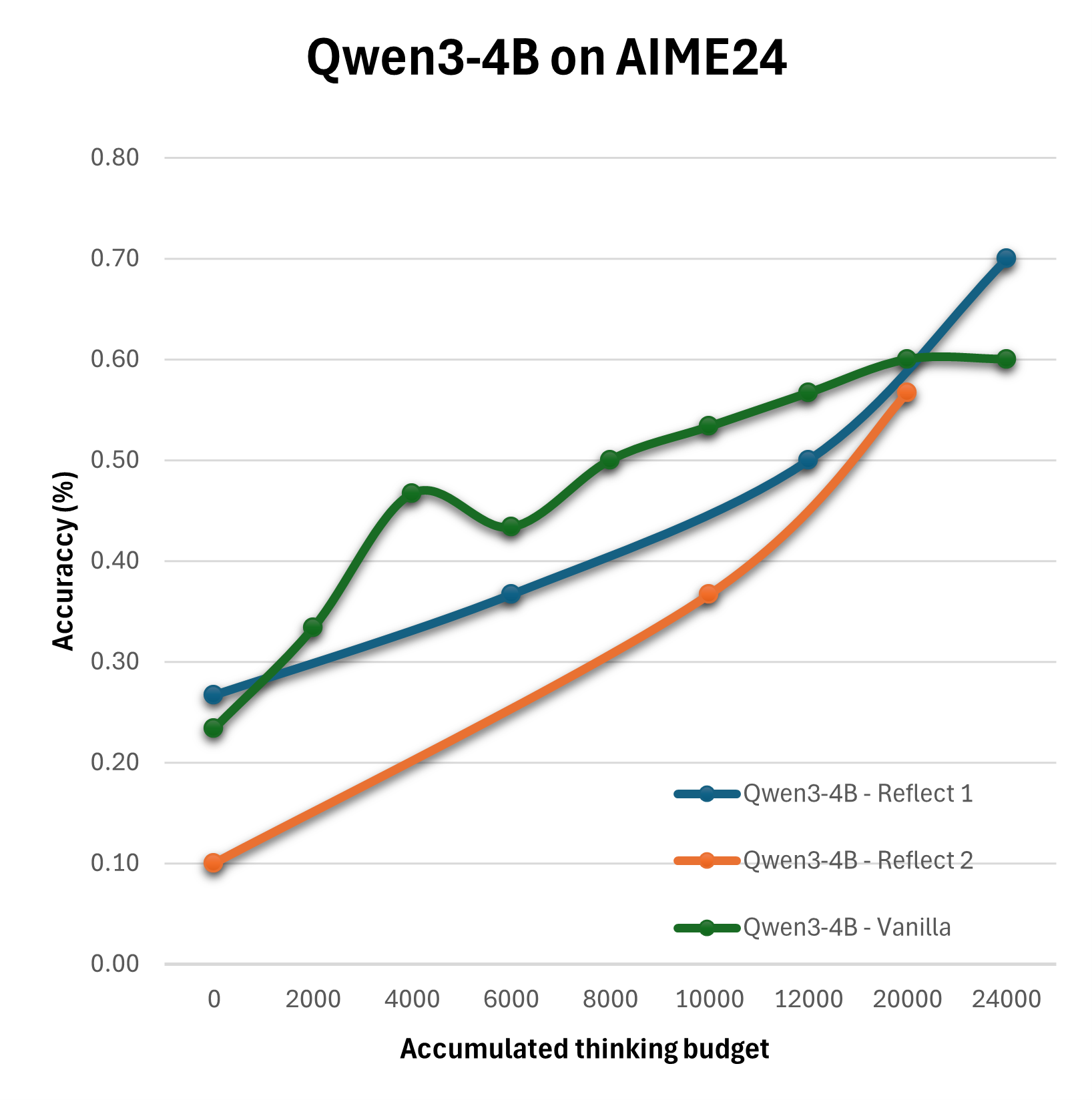}
  \end{minipage}
  \hfill
  \begin{minipage}[b]{0.30\textwidth}
    \includegraphics[width=\textwidth]{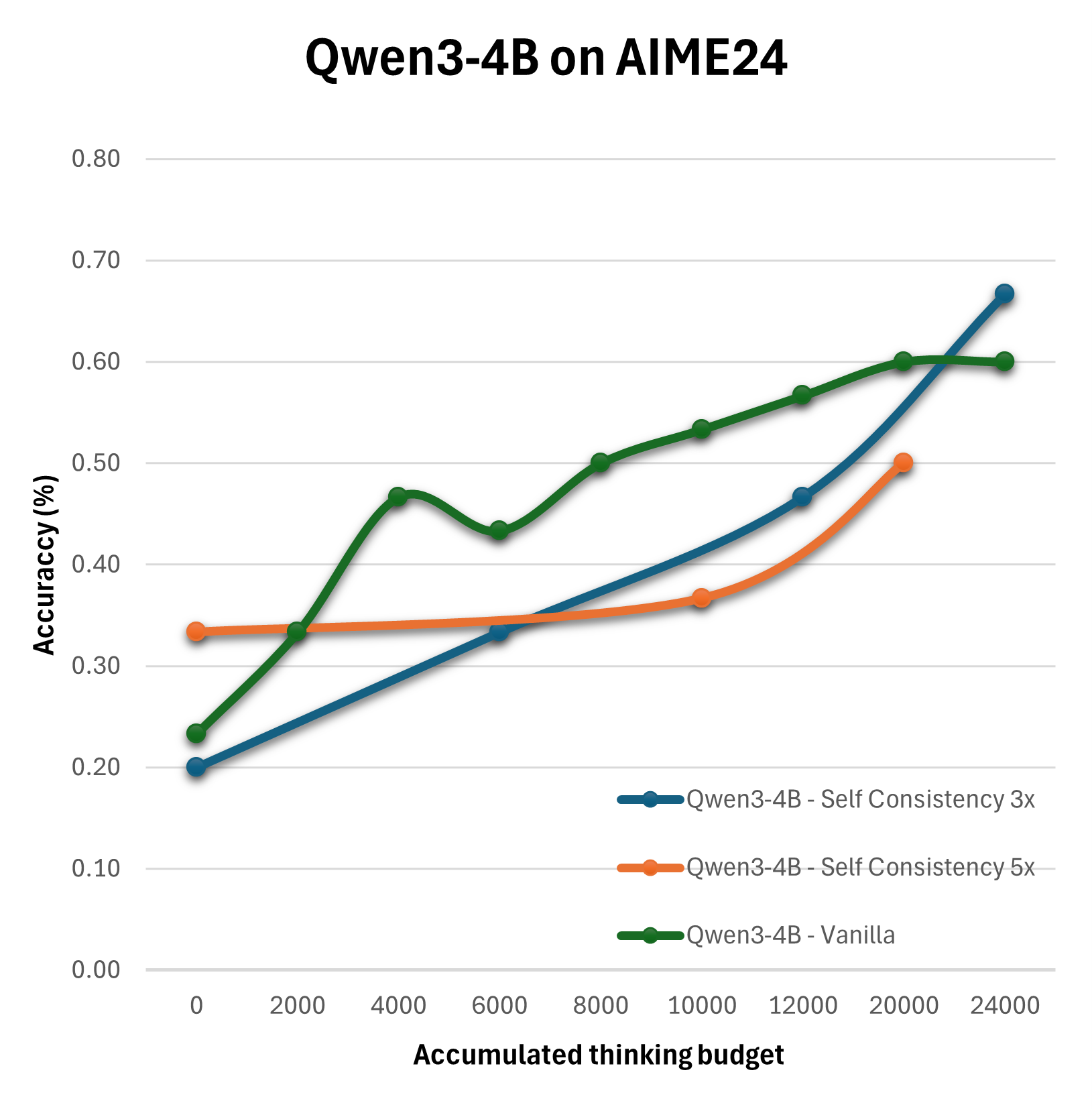}
  \end{minipage}
    \hfill
   \begin{minipage}[t]{0.30\textwidth}
    \includegraphics[width=\textwidth]{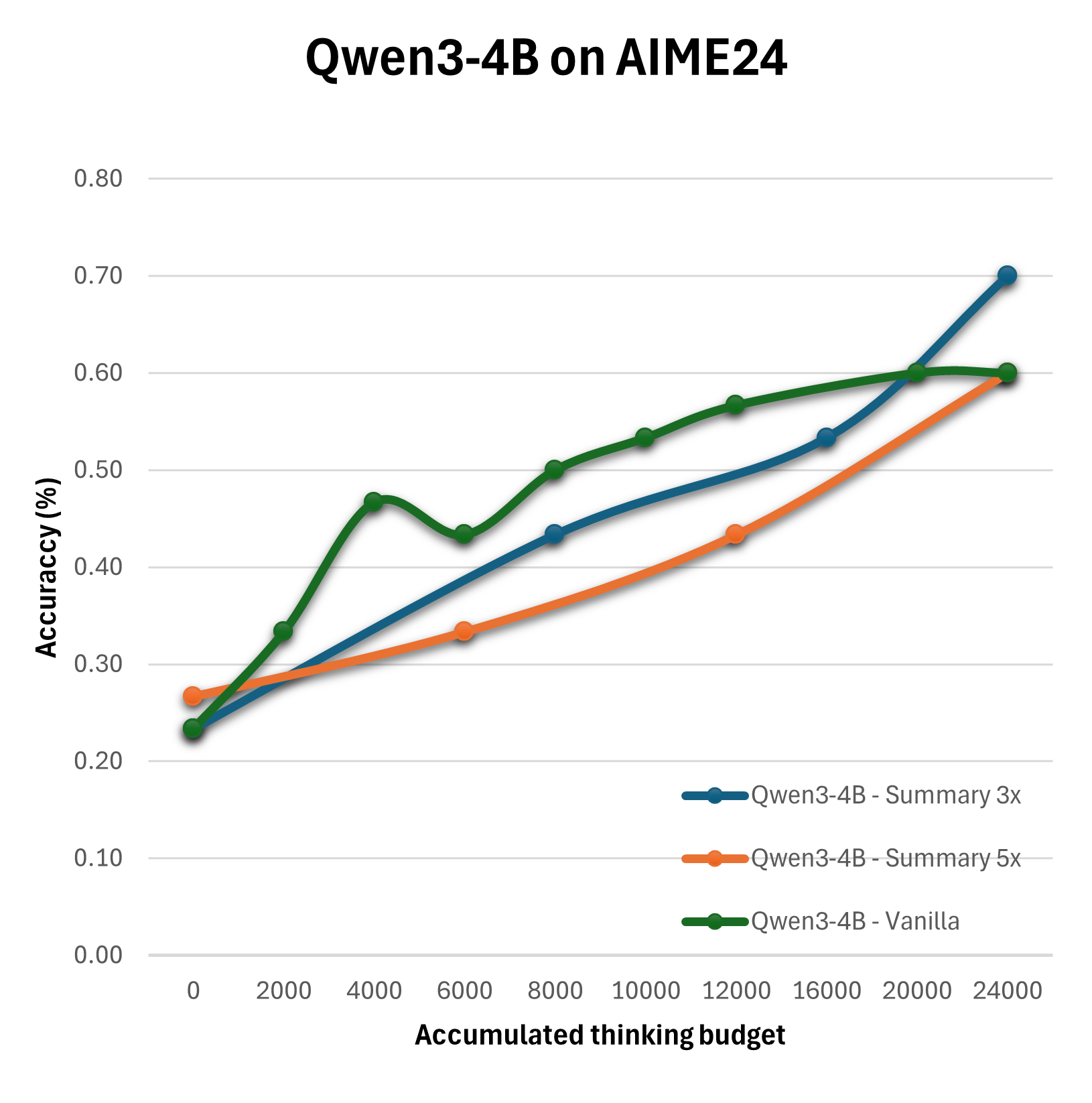}
  \end{minipage}
  \vspace{-0.2cm}
  \begin{minipage}[b]{0.30\textwidth}
    \includegraphics[width=\textwidth]{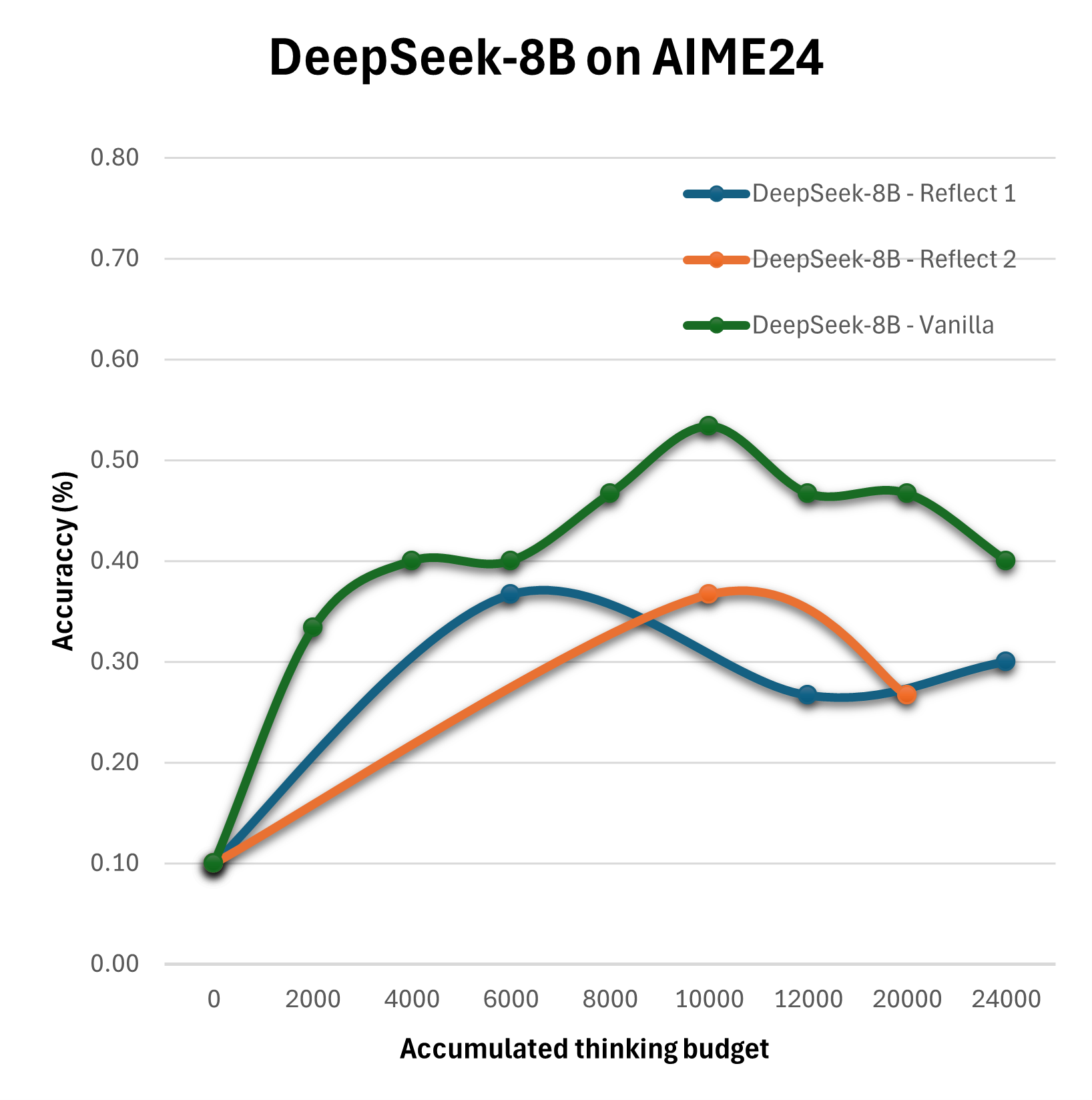}
  \end{minipage}
  \hfill
  \begin{minipage}[b]{0.30\textwidth}
    \includegraphics[width=\textwidth]{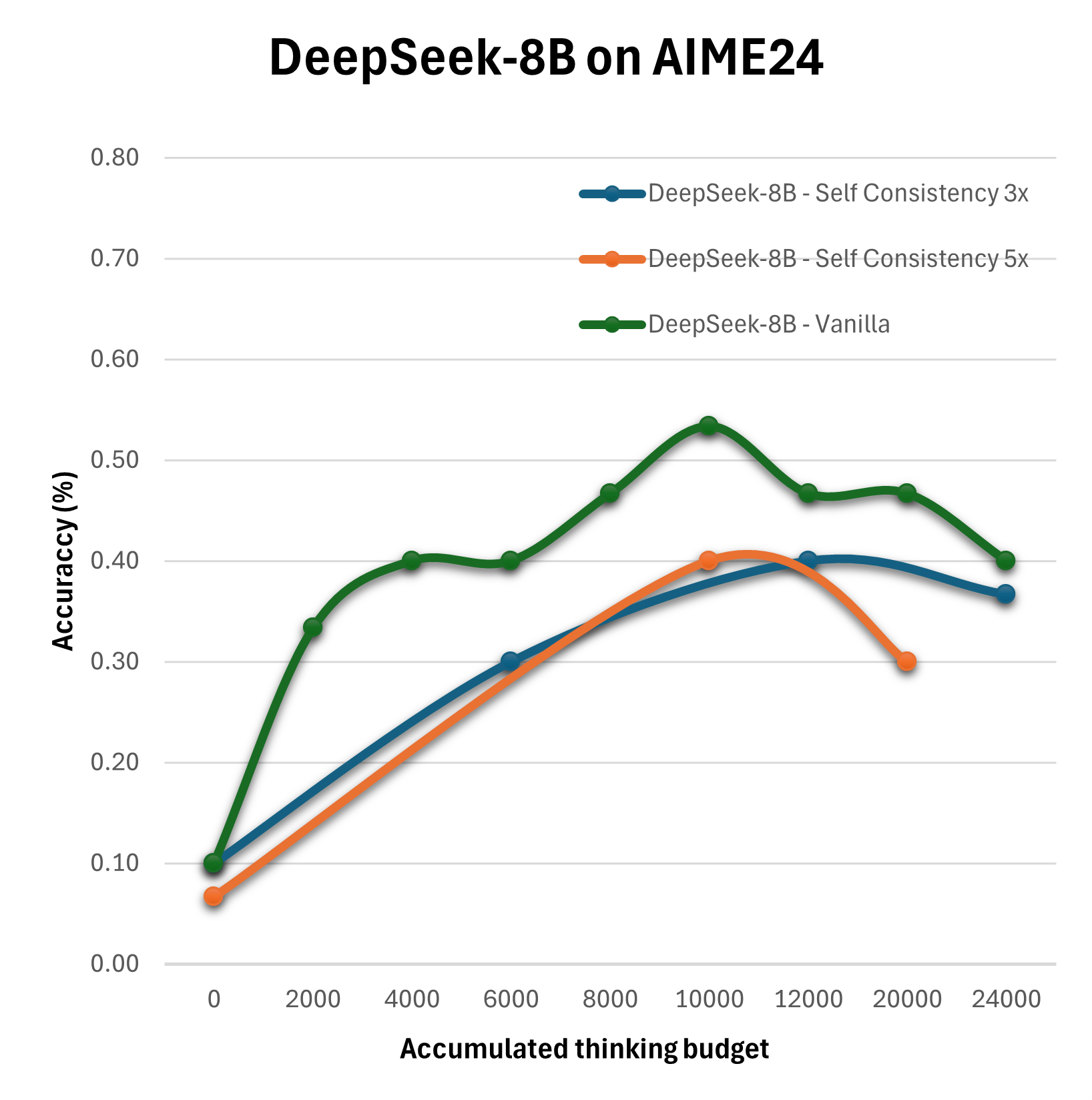}
  \end{minipage}
    \hfill
   \begin{minipage}[t]{0.30\textwidth}
    \includegraphics[width=\textwidth]{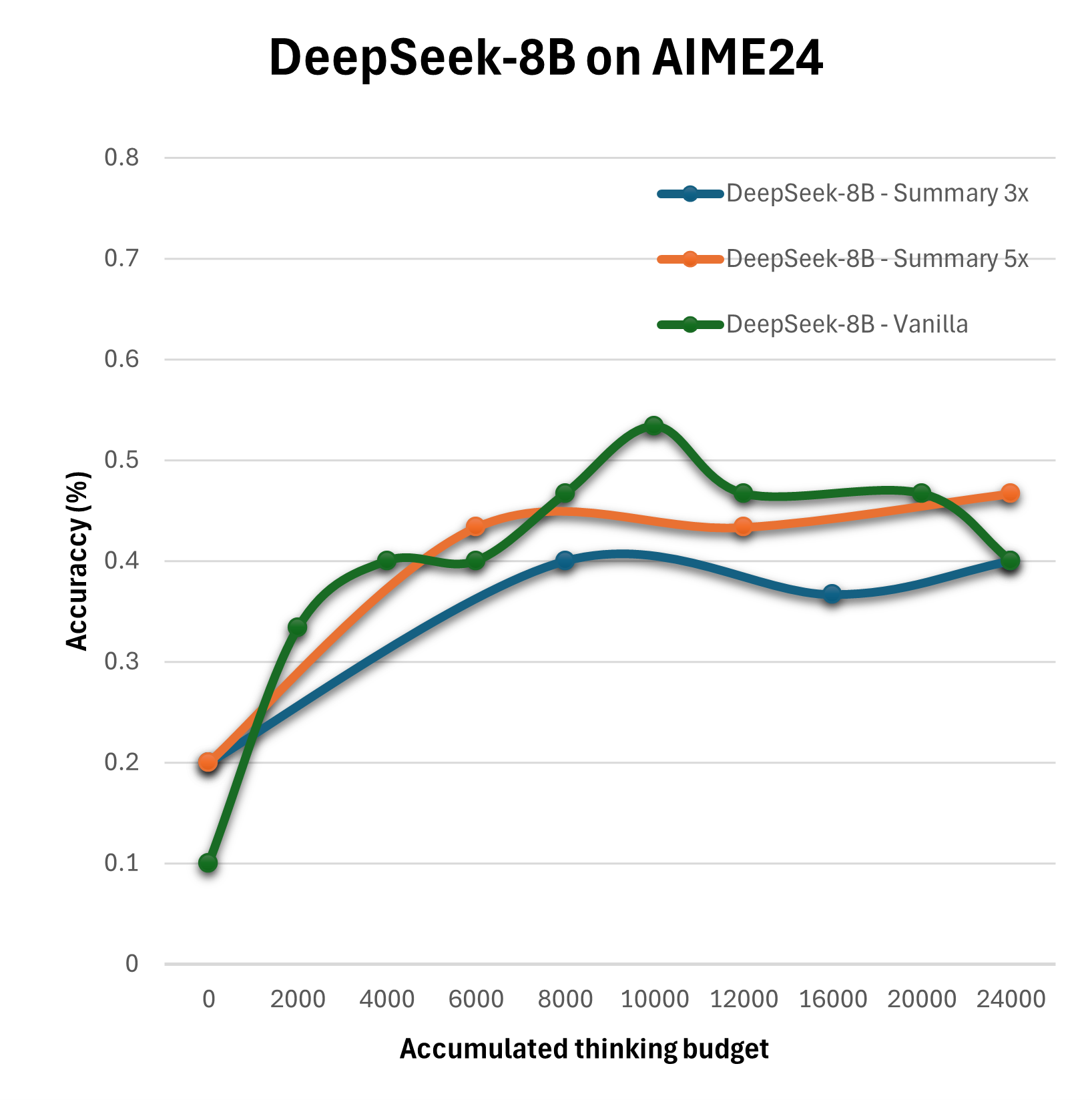}
  \end{minipage}
   \caption{Evaluation of different configurations on the AIME24 on Qwen3-8B, Qwen3-4B and DeepSeek-R1-Distill-Llama-8B.}
	\label{fig:performance}
\end{figure*}

\subsection{Analyzing the compute budget}

\citet{su2025underthinkingoverthinkingempiricalstudy} performed a thoughtful analysis of the relationship between reasoning length and answer correctness. This work shows that the length of LLM reasoning is not a good indicator of perceived difficulty, failing to increase appropriately for hard problems.
This work also finds out that LLMs often misjudge problem difficulty and shows that optimizing for shorter responses can reduce length while maintaining reasonable accuracy. Similarly, the analysis performed by 
\citet{wang-etal-2024-reasoning-token} is also an analysis paper that does not deal with reasoning LLMs but compares different reasoning approaches, single agent, multi-agent, reflection, self-consistency, etc, from the perspective of the reasoning budget. A key finding is that complex reasoning strategies often do not outperform simpler baselines due to inherent algorithmic superiority, but rather because they are allocated larger computational resources.
In a similar path, \citet{snell2024scalingllmtesttimecompute} explores how increasing the token budget of LLMs inference time can significantly boost their performance. Two strategies are investigated: "dense, process-based verifier reward models" and "adaptively updating the model's distribution over a response.". As an outcome, the authors discovered that a strategy that allocates more computational effort based on prompt difficulty is more efficient.




\section{Thinking}


Here we define what we understand by compute budget. As empirical results will show, performance often correlates with the number of thinking tokens, regardless of the amount of compute without thinking. From now on we use thinking token (and not total token output) as our metric.

\subsection{Reasoning Budget}

\paragraph{Total number of thinking tokens.} 
This is the expected number of tokens between special \textit{<think>} tags. In our experiments we do not set this value as a maximum, but we employ a forcing mechanism, following \citet{muennighoff2025s1simpletesttimescaling} that guarantees a minimum amount of tokens as defined, by inserting "Wait" tokens if the budget is not met.
\vspace{-0.25cm}
\paragraph{Number of API calls.} 
This is the amount of times the LLM is prompted to generate an answer. This value is independent from the thinking budget. For instance when we consider a 6,000 token budget, we analyze configurations where the entire budget is consumed in a single API call, or divided into three calls with 2,000 thinking tokens each. This aspect will be clarified in the Experiments section below.

\subsection{Optimizing compute}

In this section, we present different configurations aimed at answering this research question. We can visualize them in Figure \ref{fig:reasoning_models}. Each configuration is tested with varying levels of the thinking process, including configurations with no thinking at all:
\paragraph{Vanilla.} 
The model is directly prompted with a question and returns an answer without any additional reasoning steps.
\vspace{-0.27cm}
\paragraph{Self-consistency.} 
The model is prompted multiple times with temperature set to 1.0. The final answer is selected based on majority voting among the generated responses. We experimented with 3x and 5x consistency runs.
\vspace{-0.27cm}
\paragraph{Summary.} 
Similar to Self-Consistency, but instead of relying on majority voting, an additional call to the LLM is made to summarize and consolidate all attempted answers into a single, unified response. We also experimented with 3x and 5x consistency runs.
\vspace{-0.27cm}
\paragraph{Reflection.} 
Inspired by the approach proposed by \citet{madaan2023selfrefineiterativerefinementselffeedback}. This method involves generating an initial answer and then prompting the model to provide feedback on it. The original answer is subsequently refined on the basis of this feedback. We allowed one and two steps of reflection.


\section{Experiments}

For our experiments we chose three thinking models: Qwen3-8B recently introduced by \citet{qwen3} and its smaller version Qwen3-4B; and DeepSeek-R1-Distill-Llama-8B, from \citet{deepseekai2025deepseekr1incentivizingreasoningcapability}. We use the AIME24 datasets as implemented in \citet{muennighoff2025s1simpletesttimescaling}. 
In our experiments, we evaluated seven different configurations of reasoning strategies to assess their performance relative to computational cost. 
We varied the total computational budget between configurations, ranging from zero (no thinking) to 24,000 reasoning tokens.
In Figure \ref{fig:performance}, we compare model performance across the configurations described above as a function of the compute budget. We compare the Vanilla configuration individually against Reflect, Self-Consistency and Summary on the three selected models. Qwen3-8B being the strongest reasoning model seems to leverage the most of the thinking budget. Vanilla does fluctuate across the budgets, and show some signs of plateau across the board with larger allowances. This can be also seen on Figure \ref{fig:performanceAIME25} located in the Appendix. The other two weaker models, Qwen3-4B and DeepSeek-8B, struggle to fully benefit from the configurations lagging behind on most experiments against Vanilla. The Self-Consistency strategy struggle to provide any improvement, specially on those two models. This may be due to the extraction process. On this configuration, individual results are extracted via regular expressions and counted. Formatting issues might be problematic in this step, while not being an issue in the remaining configurations.
Finally we refer to the Summary strategy, which seems to be the strongest configuration overall, clearly surpassing the performance of Vanilla on Qwen3-8B and providing competitive results on the other two models.

\subsection{Zeroing the thinking budget}

Table \ref{tab:no-think} presents empirical results of experiments conducted with thinking explicitly deactivated, that is, the thinking budget was set to zero across all configurations. Under this constraint, each configuration was limited to a fixed maximum number of tokens, ensuring a fair comparison without the influence of variable reasoning depth. Consistent with findings from earlier experiments, performing two reflection steps does not yield additional benefits over a single reflection. The same applies to Self-Consistency and Summary. Increasing the amount of parallel traces is not bringing further benefits.
Among all tested configurations, Summary emerges as the top performer. It enables improved results even in the weakest model of the three. Despite the absence of an extended reasoning phase, this setup still benefits from generating multiple diverse outputs and concluding the most consistent response. This indicates that even modest ensemble strategies can significantly enhance answer quality.


\begin{table}[t]
  \small
    \renewcommand{\arraystretch}{0.85}
    \setlength{\tabcolsep}{3pt}
  \centering
  \begin{tabular}{l | c  c c }
    \toprule
    \textbf{Configuration} &  Qwen3-8B &		Qwen3-4B &	DeepSeek-8B	 \\
    \midrule
    Vanilla	                &  23.33\%  & 23.33\%	& 10.00\%	\\
    \midrule
    Self Consistency 3x	    & 33.33\% & 20.00\%	& 10.00\%	\\
    Reflect 1 	            &  26.67\% & 26.67\% & 10.00\%	\\
    \midrule    
    Summary 3x              & 33.33\% &	23.33\% & 20.00\%   \\
    \midrule
    Self Consistency 5x     &  30.00\% & 33.33\% & 6.67\% \\	
    Reflect 2	            &  26.67\% &	10.00\% & 10.00\% \\
    \midrule
    Summary 5x             &  30.00\% & 26.67\% & 20.00\% \\
    \bottomrule
  \end{tabular}
  \caption{\label{tab:no-think}
    Performance when thinking is deactivated. 
  }
\end{table}

\section{Conclusion}

In this paper we provide an initial examination of the behavior of several reasoning strategies while varying their thinking budget. 
We confirmed that increasing the compute budget leads to better performance, but simpler configurations reach a performance ceiling that can be overcome by more sophisticated strategies. Overall the \textit{Summary} strategy seems to be the best-performing.

\newpage

\section*{Limitations} 
Although our analysis shows interesting findings, we carried out experiments on a limited set of datasets and LLMs. A larger sample both in terms of models and benchmarks is necessary to fully validate our current findings.

\section*{AI Assistants} 
The OpenAI service of \href{https://chatgpt.com/}{ChatGPT} was used to polishing authors writing. 

\bibliography{custom}

\appendix

\section{Experiment configuration}

We followed the framework introduced by \citet{muennighoff2025s1simpletesttimescaling}, \url{http://github.com/simplescaling/s1}, which is a modified version of Language Model Evaluation Harness from EleutherAI \cite{eval-harness}. \\
Call:
\begin{verbatim}
OPENAI_API_KEY=<YOUR_OPENAI_KEY>
PROCESSOR=gpt-4o-mini 
lm_eval --model vllm \
  --model_args <model-params> \
  --tasks aime24_figures,aime24_nofigures \
  --batch_size auto \
  --apply_chat_template 
\end{verbatim}

\noindent All configurations were implemented using LangChain \cite{chase2024langchain} framework.
\noindent Temperature on inference was set to zero except on Self-Consistency and Summary, where we use 1.0 instead.

\section{Judge LLM}
This configuration uses an extra call to the LLM, without thinking to check if the answer provided is correct or another try is needed. Importantly, this judging step does not involve further reasoning; it merely evaluates whether the answer is acceptable or whether a new attempt is required. \\
Three configurations were considered depicted in Figure \ref{fig:reasoning_models2}: 

\begin{figure}[t!]
	\centering
	\includegraphics[width=0.9\linewidth]{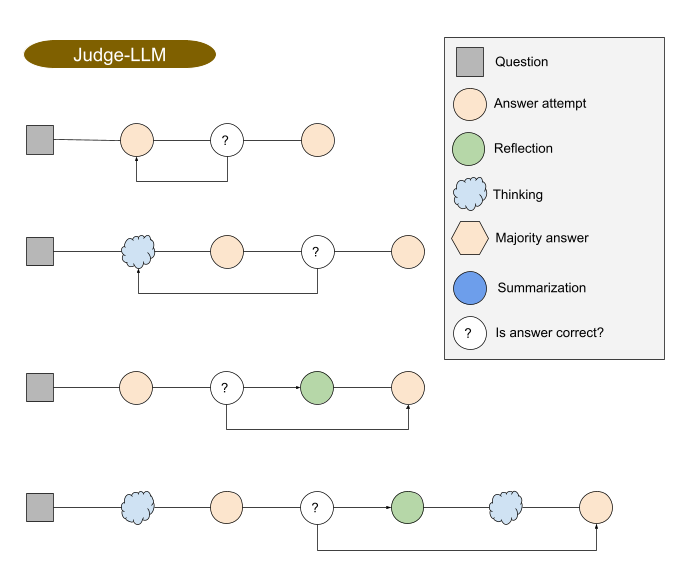}
	\caption{Visualization of reasoning strategies across Judge LLM configurations.}
	\label{fig:reasoning_models2}
\end{figure}

\paragraph{Judge-LLM.} 
After producing an answer, the model is prompted to evaluate whether the response is likely correct. If deemed incorrect, the model is given another opportunity to respond. 
\vspace{-0.25cm}
\paragraph{Judge-LLM w/history} 
Similar to Judge-LLM, but in this variant, the model’s second attempt is informed by both the initial answer and the negative feedback, which are included in the prompt.
\vspace{-0.25cm}
\paragraph{Judge+Reflection.} 
A hybrid of the Judge-LLM and Self-Reflection methods. If the initial answer is judged incorrect, the model is prompted to generate feedback and then refine its response accordingly.

\begin{table}[h!]

  \centering
      \renewcommand{\arraystretch}{1.0}
     \setlength{\tabcolsep}{3pt}
  \begin{tabular}{l c  c  | c c }
      \toprule
    \textbf{Configuration}  & Think & Budget & Qwen3-8B  \\
    \midrule
    \multirow{3}{*}{Vanilla}    &            & -     & 23.33\%  \\
                            & \checkmark & 2000 & 46.67\%  \\
                            & \checkmark & 4000 &  46.67\%  \\
    \midrule
    \multirow{3}{*}{Judge-LLM w/o history} &          & -       & 43.33\% \\
                             & \checkmark & 2000     & 60.00\% \\
                             & \checkmark & 4000    & 66.67\% \\

    \midrule
    \multirow{3}{*}{Judge-LLM with history} &          & -     & 46.67\% \\
                             & \checkmark & 2000    & 63.33\% \\
                             & \checkmark & 4000     & 63.33\%  \\
    \midrule
    \multirow{3}{*}{Judge+Reflection} &          & - & 40.00\% \\
                             & \checkmark & 2000    &  46.67\% \\
                             & \checkmark & 4000    &  40.00\% \\
        \bottomrule
  \end{tabular}
  \caption{\label{tab:judgellm}
    Comparison between vanilla configuration and Judge LLM.
  }
\end{table}

Experiments were carried out using Qwen3-8B on AIME24 dataset. Table \ref{tab:judgellm} shows that the inclusion of a Judge LLM appears to be beneficial across a range of configurations and compute budgets. For the first configuration this results is curious since if model values the answer as incorrect, the answer is removed from history, and the model is just answering from scratch. 
Comparing the first two Judge-LLM configurations, given the mixed results remains unclear whether keeping previous answer is beneficial or not. Meanwhile Judge+Reflection struggles to improve over the Judge-LLM strategy. The result is unclear but an analysis on the results show most of the times the first answer is assess as valid, neglecting the potential improvement from the reflection feedback.

 \newpage

\section{Prompts used}

Here we show the prompts used on the configurations mentioned in the main body of the paper:

\textbf{Vanilla.} and \textbf{Self-consistency.}
\begin{assistantbox}
"You are a helpful assistant tasked to answer user's questions"
\end{assistantbox}

\textbf{Summary.} 
\begin{assistantbox}
"You are a helpful assistant tasked to answer user's questions"
\end{assistantbox}

\begin{summarizationbox}
Given this question\\
${q}$\\
we have a set of potential answers:\\
${docs}$\\
Take these and distill it into a final, consolidated answer.
\end{summarizationbox}

\textbf{Reflection.} 
\begin{assistantbox}
"You are a helpful assistant tasked to solve puzzles. " \\
"If the user provides feedback, respond again a revised version of your previous attempts. " \\
"DO NOT thank for the feedback. Pretend it's the first time you are answering the question. "
\end{assistantbox}
\begin{reflectionbox}
"You are a helpful assistant able to grade answers. " \\
"Check correctness of the user's answer. " \\
"Provide feedback and detailed recommendations, including requests for length, depth, etc.,"
\end{reflectionbox}

\begin{table*}[t]
  \centering
  \small
  \renewcommand{\arraystretch}{1.3}
  \setlength{\tabcolsep}{5pt}
  \begin{tabular}{l c r c r | c c c c c c }
    \toprule
    \multirow{2}{*}{\textbf{Configuration}} & \multirow{2}{*}{Thinking} & \multirow{2}{*}{Budget} & \multirow{2}{*}{Calls} & \multirow{2}{*}{Total} & \multicolumn{2}{c}{Qwen3-8B} & \multicolumn{2}{c}{Qwen3-4B} & \multicolumn{2}{c}{DeepSeek-8B} \\
    &&&&& AIME24 & AIME25 & AIME24 & AIME25 & AIME24 & AIME25\\
    \midrule
\multirow{9}{*}{Vanilla} & & 0    & 1 & 0     & 23.33\% & 20.00\% & 23.33\% & 20.00\% & 10.00\% & 6.67\% \\
& \checkmark & 2000 & 1 & 2000  & 46.67\% & 46.67\% & 33.33\% & 40.00\% & 33.33\% & 33.33\% \\
& \checkmark & 4000 & 1 & 4000  & 46.67\% & 46.67\% & 46.67\% & 33.33\% & 40.00\% & 33.33\% \\
& \checkmark & 6000 & 1 & 6000  & 63.33\% & 46.67\% & 43.33\% & 46.67\% & 40.00\% & 40.00\% \\
& \checkmark & 8000 & 1 & 8000  & 56.67\% & 46.67\% & 50.00\% & 46.67\% & 46.67\% & 40.00\% \\
& \checkmark& 10000 & 1 & 10000 & 53.33\% & 53.33\% & 53.33\% & 46.67\% & 53.33\% & 40.00\% \\
& \checkmark & 12000 & 1 & 12000 & 66.67\% & 53.33\% & 56.67\% & 46.67\% & 46.67\% & 40.00\% \\
& \checkmark & 20000 & 1 & 20000 & 73.33\% & 53.33\% & 60.00\% & 53.33\% & 46.67\% & 40.00\% \\
& \checkmark & 24000 & 1 & 24000 & 66.67\% & 60.00\% & 60.00\% & 53.33\% & 40.00\% & 33.33\% \\
\midrule
\multirow{4}{*}{Self Consistency 3x} & & 0    & 3 & 0     & 33.33\% & 20.00\% & 20.00\% & 33.33\% & 10.00\% & 13.33\% \\
& \checkmark  & 2000 & 3 & 6000  & 53.33\% & 53.33\% & 33.33\% & 40.00\% & 30.00\% & 20.00\% \\
& \checkmark  & 4000 & 3 & 12000 & 70.00\% & 46.67\% & 46.67\% & 33.33\% & 40.00\% & 26.67\% \\
& \checkmark  & 8000 & 3 & 24000 & 63.33\% & 60.00\% & 66.67\% & 53.33\% & 36.67\% & 46.67\% \\
\midrule
\multirow{3}{*}{Self Consistency 5x} & & 0    & 5 & 0     & 30.00\% & 20.00\% & 33.33\% & 40.00\% & 6.67\%  & 13.33\% \\
& \checkmark & 2000 & 5 & 10000 & 60.00\% & 46.67\% & 36.67\% & 40.00\% & 40.00\% & 26.67\% \\
& \checkmark & 4000 & 5 & 20000 & 63.33\% & 53.33\% & 50.00\% & 40.00\% & 30.00\% & 46.67\% \\
\midrule
\multirow{4}{*}{Reflect 1} & & 0    & 3 & 0     & 26.67\% & 26.67\% & 26.67\% & 26.67\% & 10.00\% & 6.67\% \\
& \checkmark & 2000 & 3 & 6000  & 60.00\% & 46.67\% & 36.67\% & 33.33\% & 36.67\% & 20.00\% \\
& \checkmark & 4000 & 3 & 12000 & 63.33\% & 46.67\% & 50.00\% & 40.00\% & 26.67\% & 20.00\% \\
& \checkmark & 8000 & 3 & 24000 & 60.00\% & 66.67\% & 70.00\% & 46.67\% & 30.00\% & 33.33\% \\
\midrule
\multirow{3}{*}{Reflect 2} & & 0    & 5 & 0     & 26.67\% & 20.00\% & 10.00\% & 20.00\% & 10.00\% & 13.33\% \\
& \checkmark & 2000 & 5 & 10000 & 66.67\% & 40.00\% & 36.67\% & 33.33\% & 36.67\% & 20.00\% \\
& \checkmark & 4000 & 5 & 20000 & 66.67\% & 53.33\% & 56.67\% & 46.67\% & 26.67\% & 26.67\% \\
\midrule
\multirow{4}{*}{Summary 3x} & & 0    & 4 & 0     & 33.33\% & 33.33\% & 23.33\% & 33.33\% & 20.00\% & 13.33\% \\
& \checkmark & 2000 & 4 & 8000  & 60.00\% & 53.33\% & 43.33\% & 33.33\% & 40.00\% & 20.00\% \\
& \checkmark & 4000 & 4 & 16000 & 70.00\% & 53.33\% & 53.33\% & 33.33\% & 36.67\% & 33.33\% \\
& \checkmark & 6000 & 4 & 24000 & 80.00\% & 73.33\% & 70.00\% & 53.33\% & 40.00\% & 33.33\% \\
\midrule
\multirow{4}{*}{Summary 5x} & & 0    & 6 & 0     & 30.00\% & 26.67\% & 26.67\% & 26.67\% & 20.00\% & 6.67\% \\
& \checkmark & 1000 & 6 & 6000  & 56.67\% & 53.33\% & 33.33\% & 40.00\% & 43.33\% & 33.33\% \\
& \checkmark & 2000 & 6 & 12000 & 53.33\% & 60.00\% & 43.33\% & 53.33\% & 43.33\% & 33.33\% \\
& \checkmark & 4000 & 6 & 24000 & 70.00\% & 53.33\% & 60.00\% & 53.33\% & 46.67\% & 46.67\% \\
\bottomrule
  \end{tabular}
  \caption{\label{citation-guide}
   Complete experiments table
  }
\end{table*}

\newpage

\begin{figure*}[t!]
	\centering
  \begin{minipage}[b]{0.30\textwidth}
    \includegraphics[width=\textwidth]{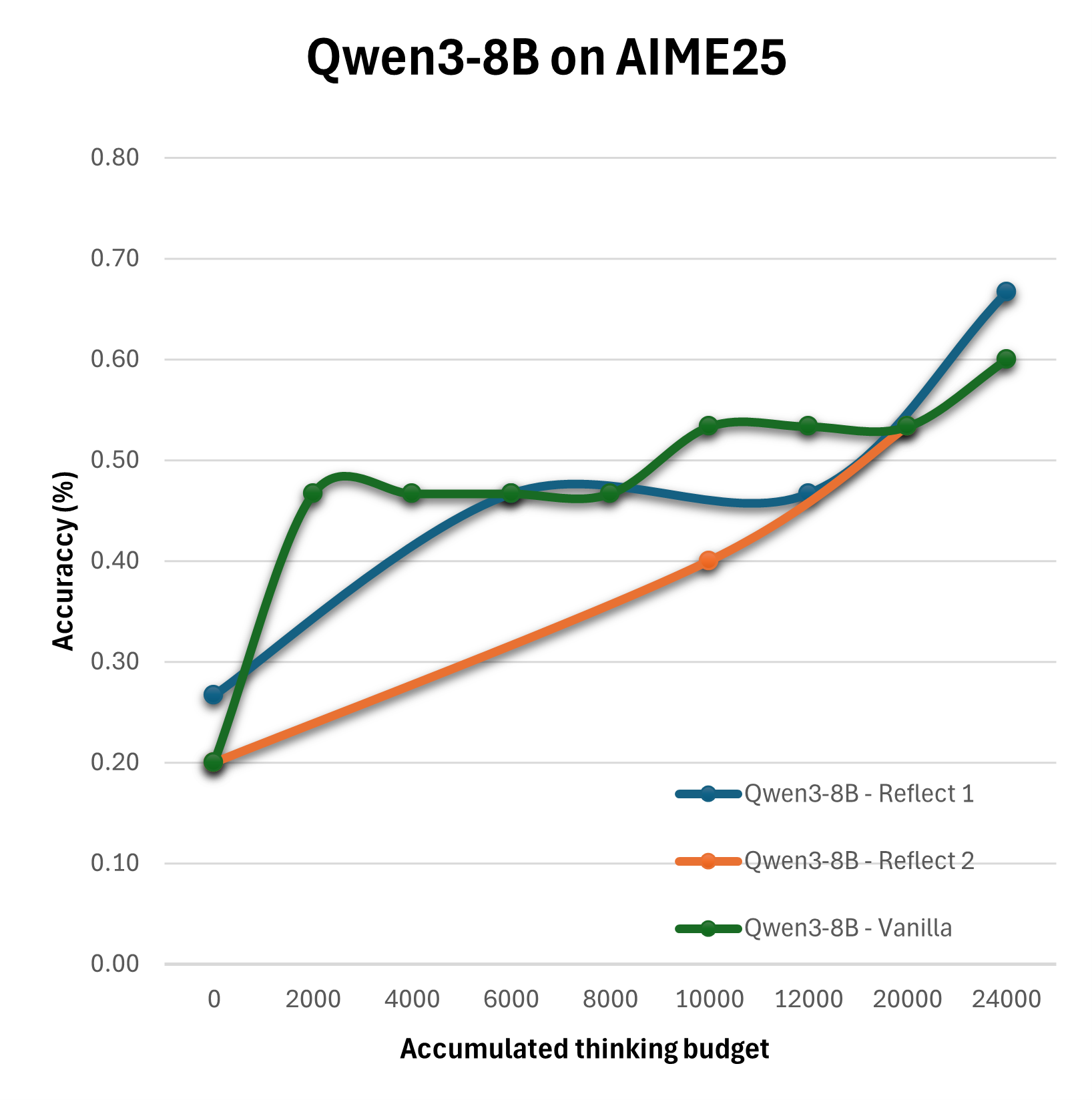}
  \end{minipage}
  \hfill
  \begin{minipage}[b]{0.30\textwidth}
    \includegraphics[width=\textwidth]{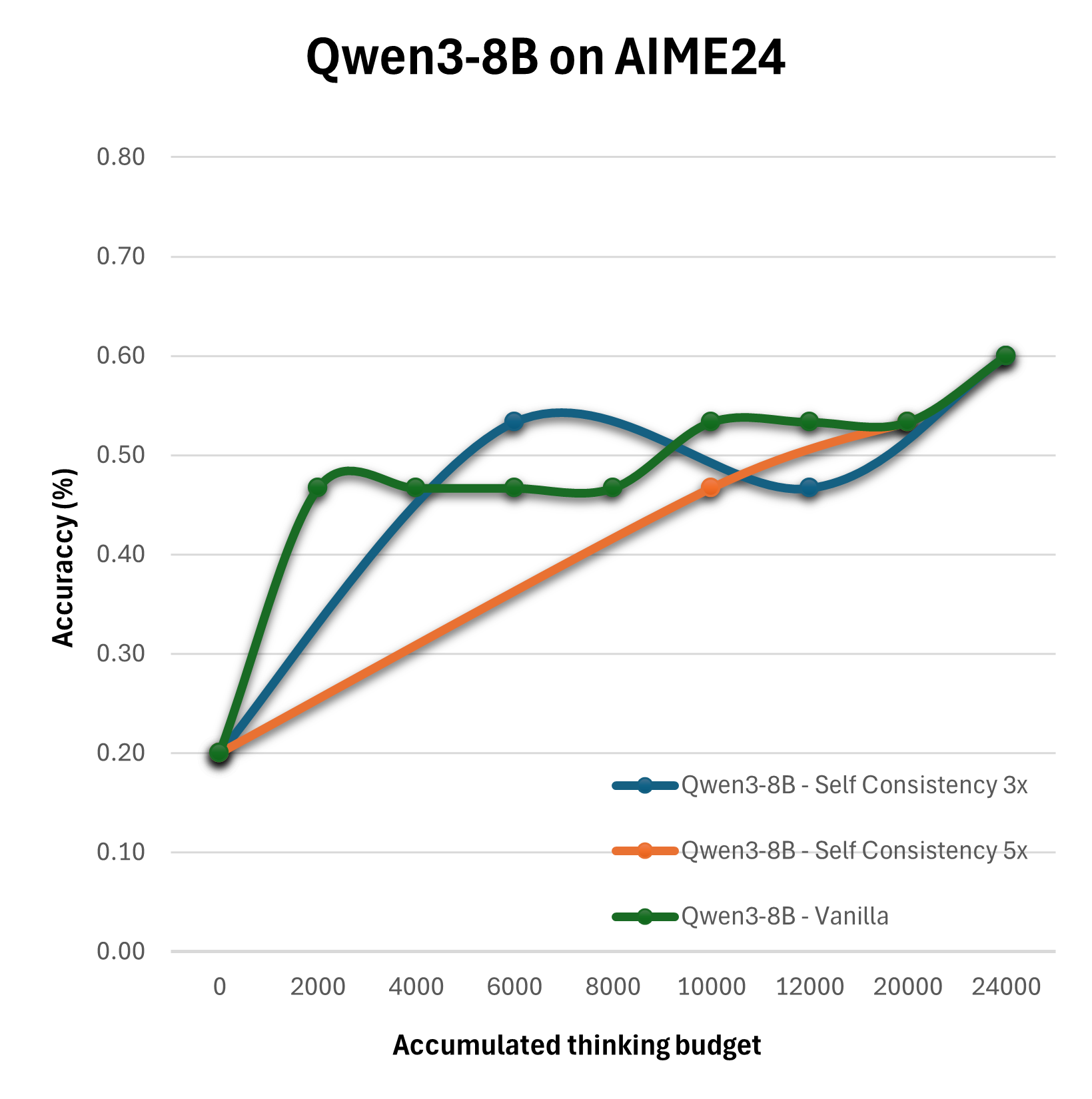}
  \end{minipage}
    \hfill
   \begin{minipage}[t]{0.30\textwidth}
    \includegraphics[width=\textwidth]{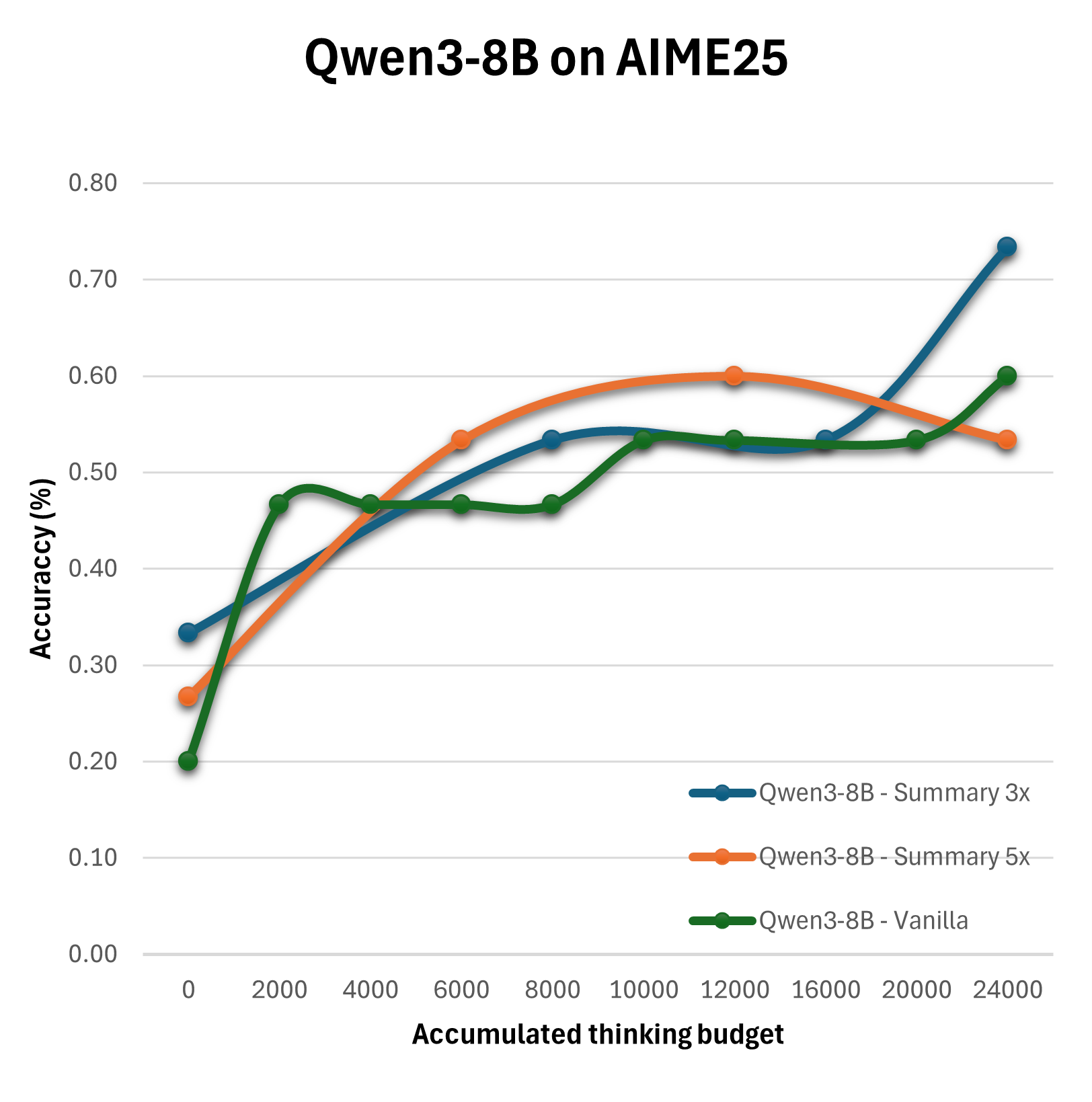}
  \end{minipage}
  \vspace{0.2cm}
  \begin{minipage}[b]{0.30\textwidth}
    \includegraphics[width=\textwidth]{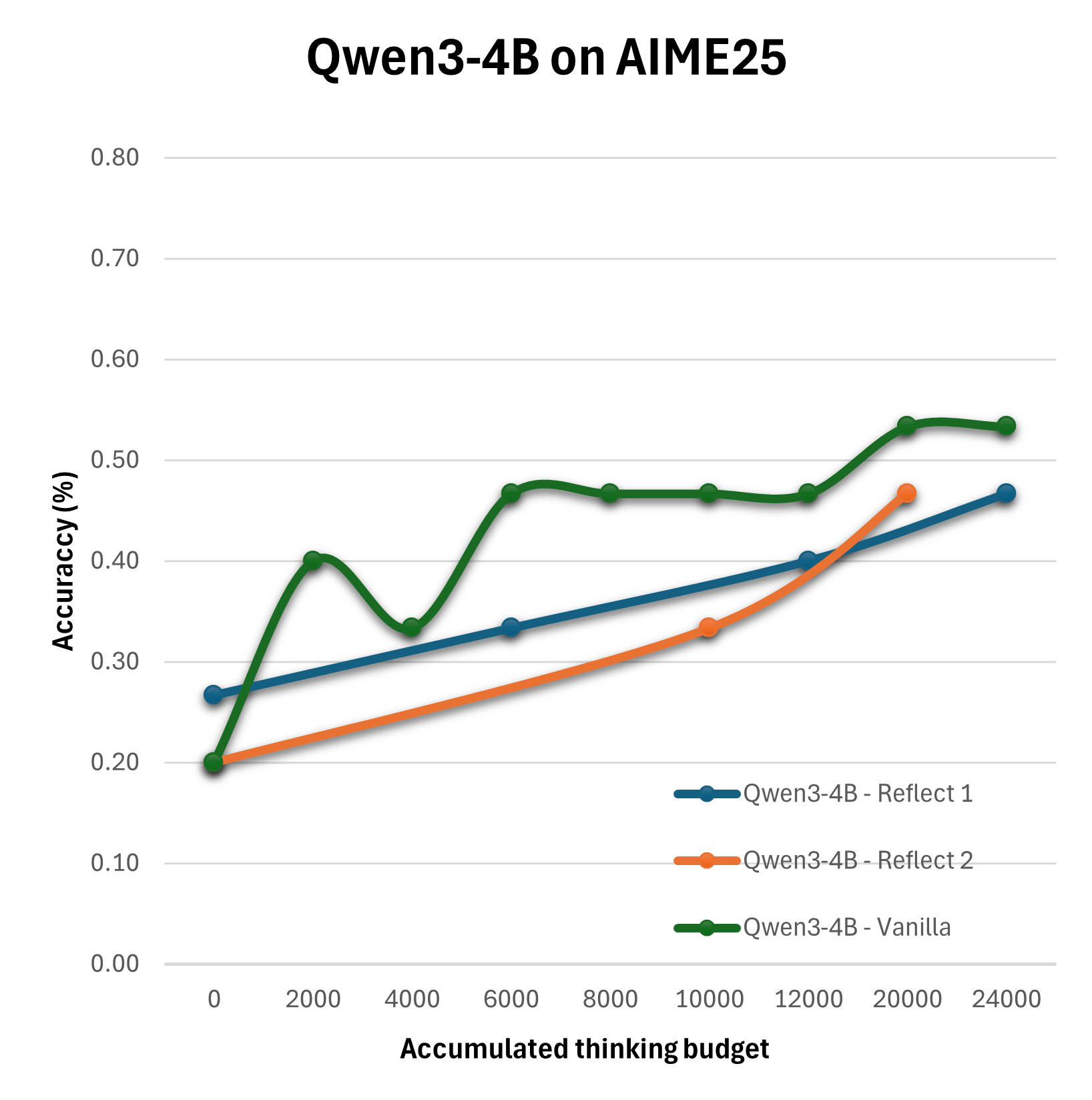}
  \end{minipage}
  \hfill
  \begin{minipage}[b]{0.30\textwidth}
    \includegraphics[width=\textwidth]{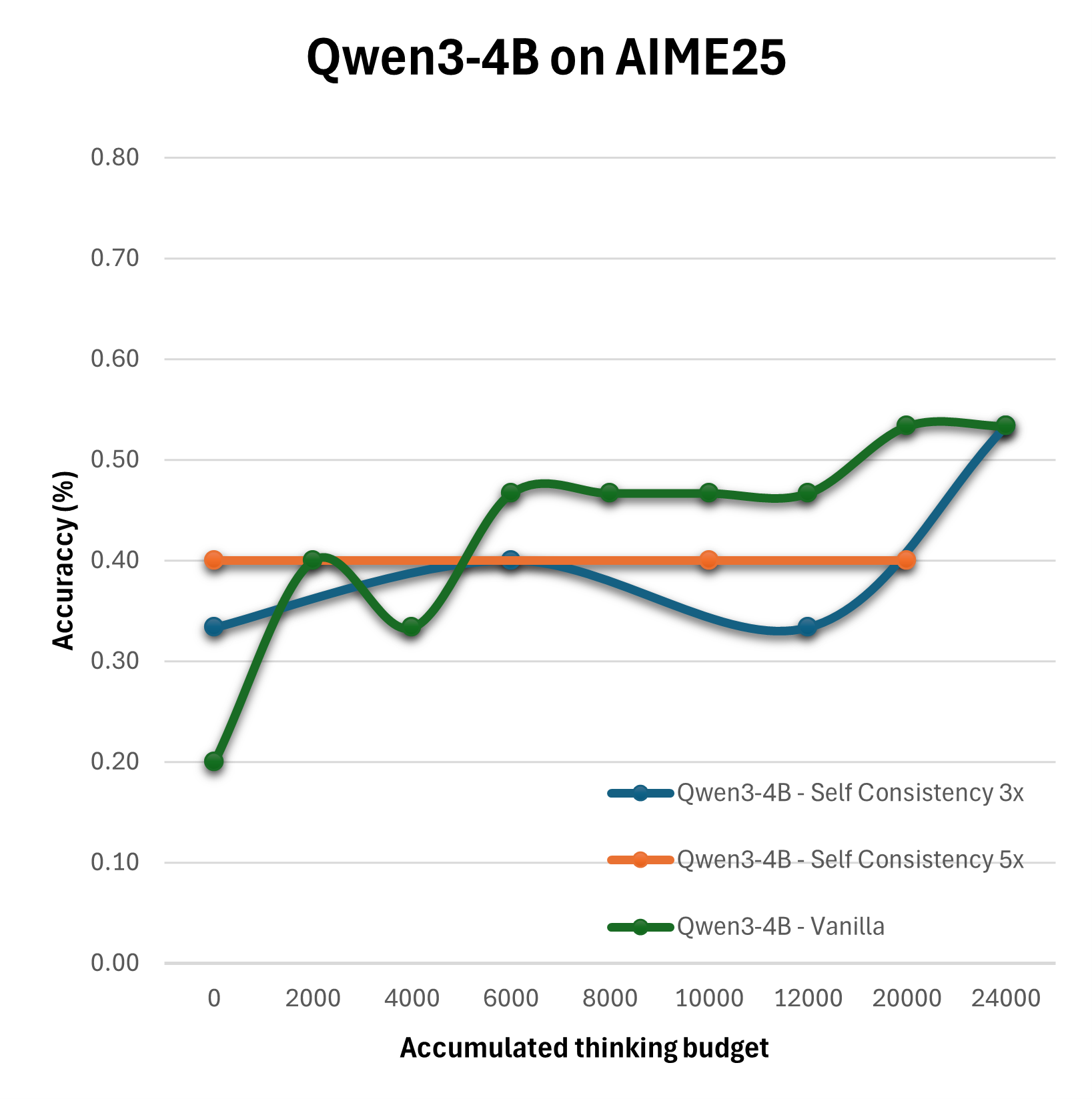}
  \end{minipage}
    \hfill
   \begin{minipage}[t]{0.30\textwidth}
    \includegraphics[width=\textwidth]{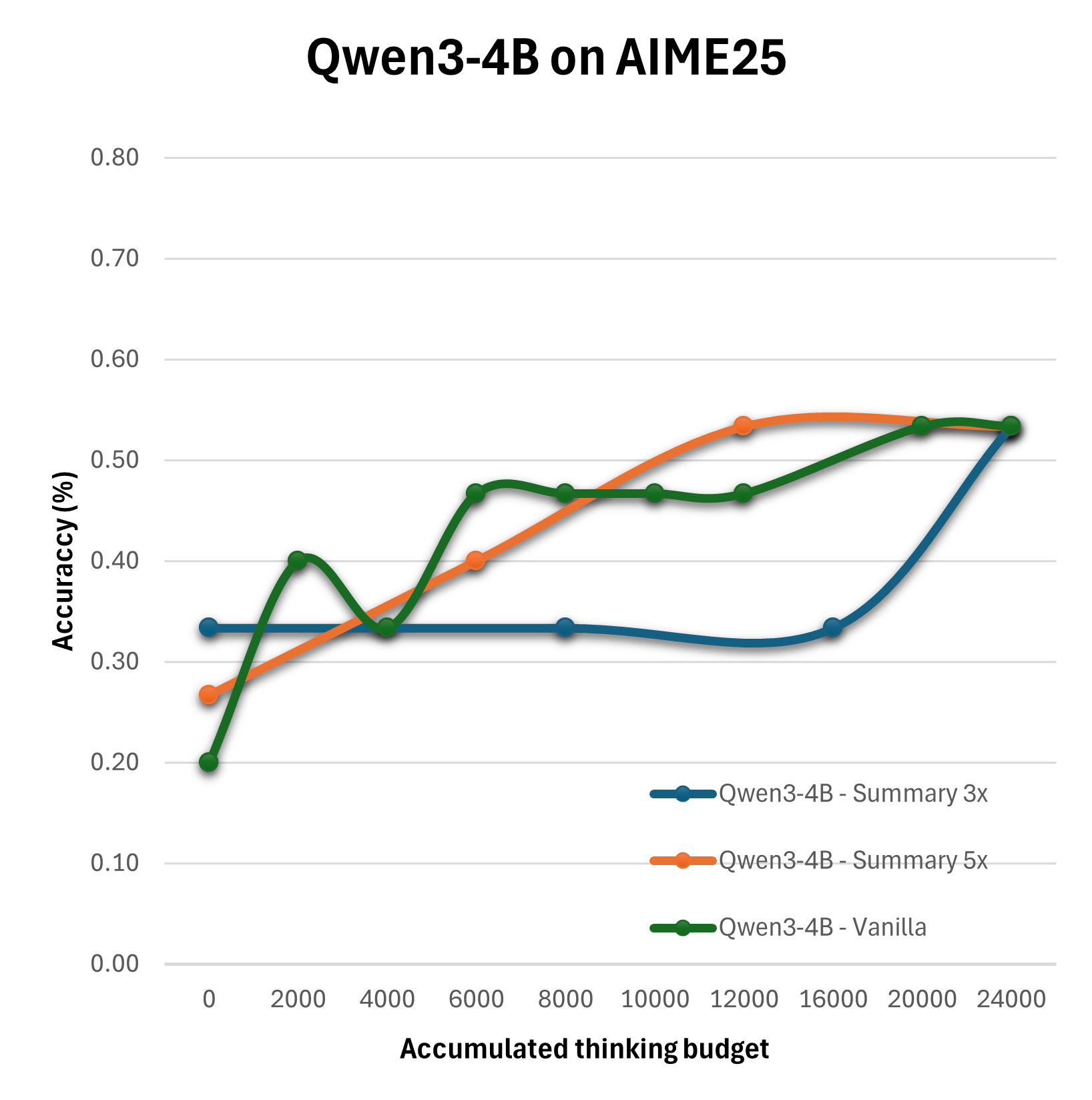}
  \end{minipage}
  \vspace{0.2cm}
  \begin{minipage}[b]{0.30\textwidth}
    \includegraphics[width=\textwidth]{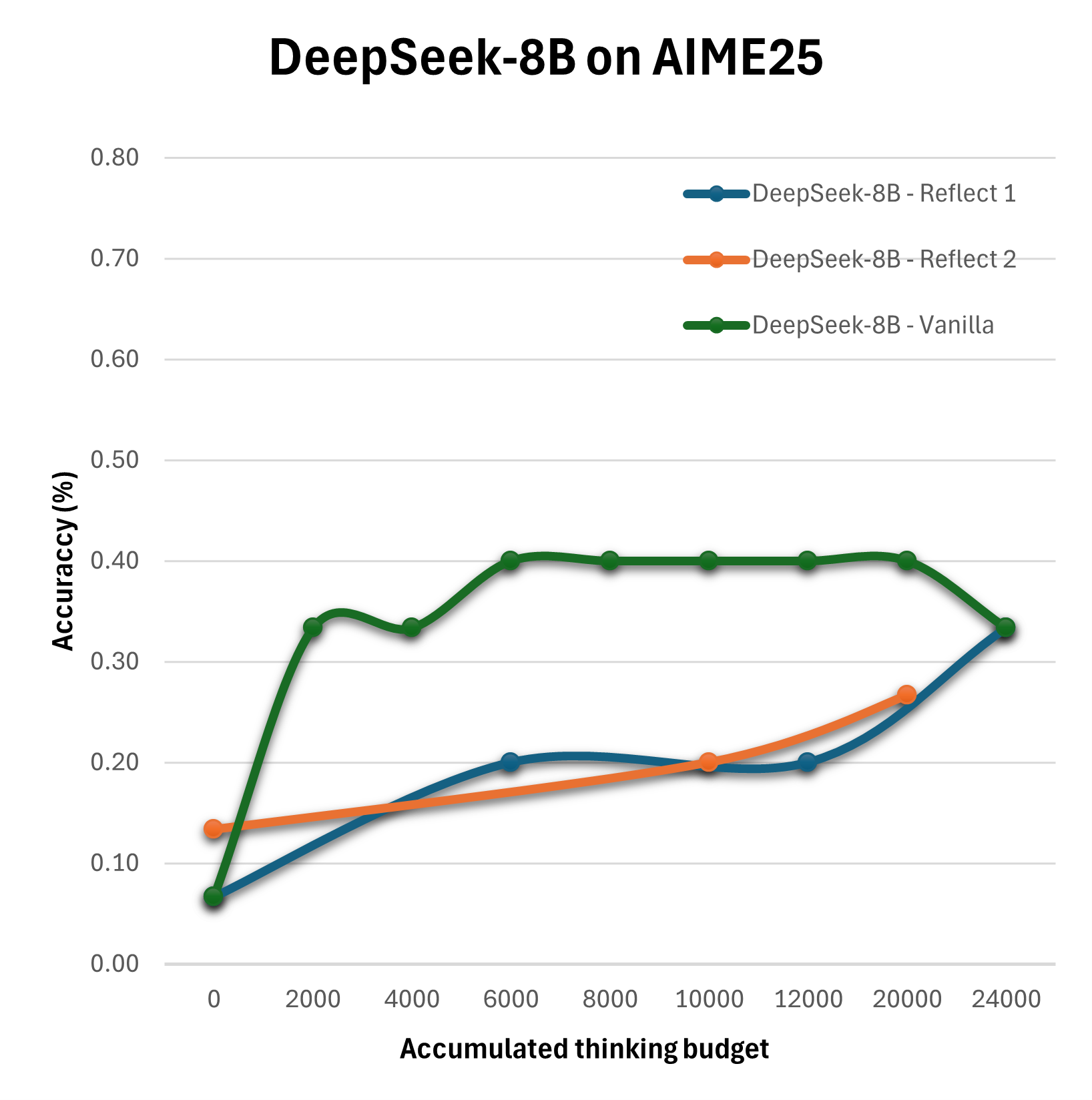}
  \end{minipage}
  \hfill
  \begin{minipage}[b]{0.30\textwidth}
    \includegraphics[width=\textwidth]{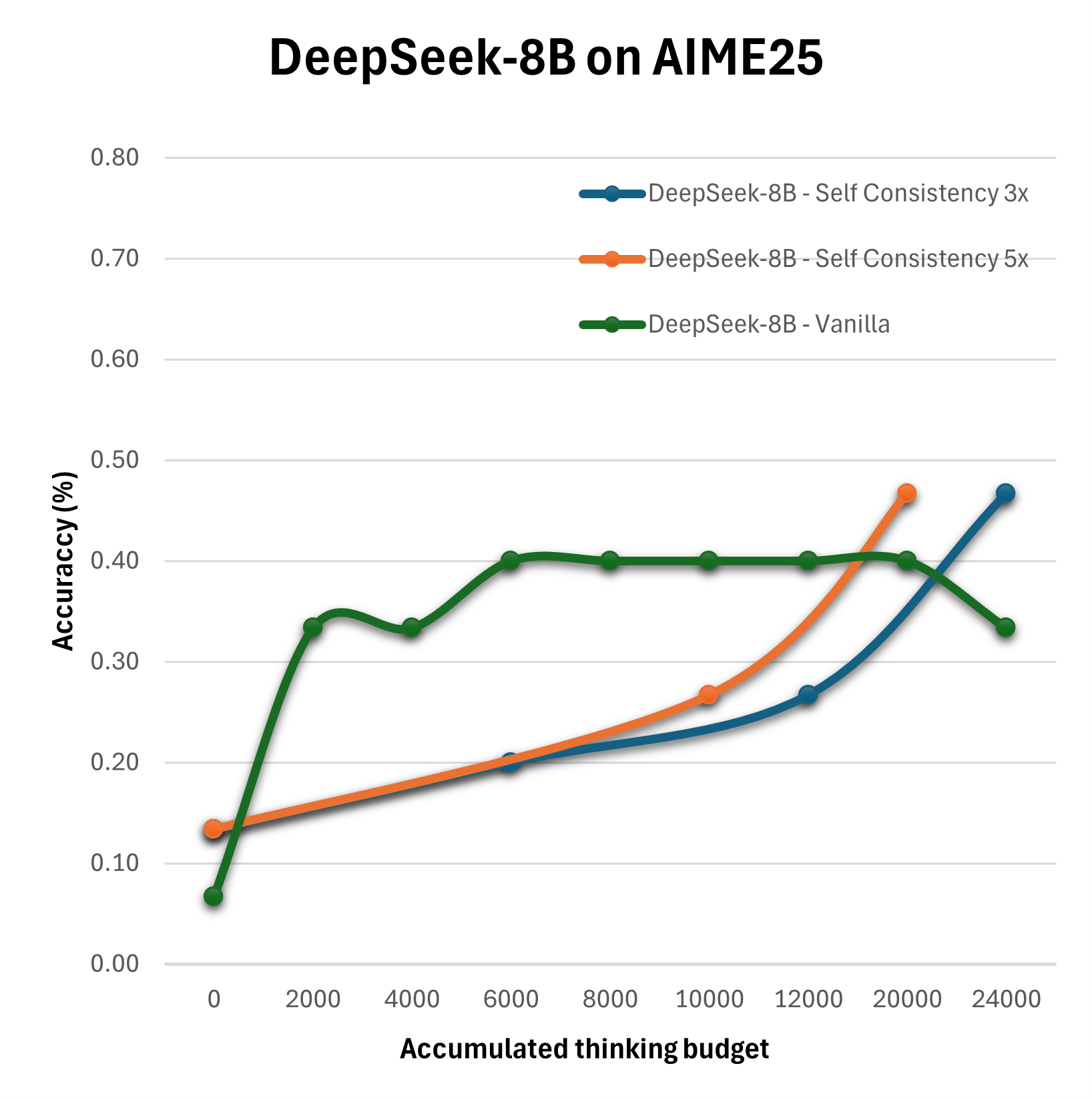}
  \end{minipage}
    \hfill
   \begin{minipage}[t]{0.30\textwidth}
    \includegraphics[width=\textwidth]{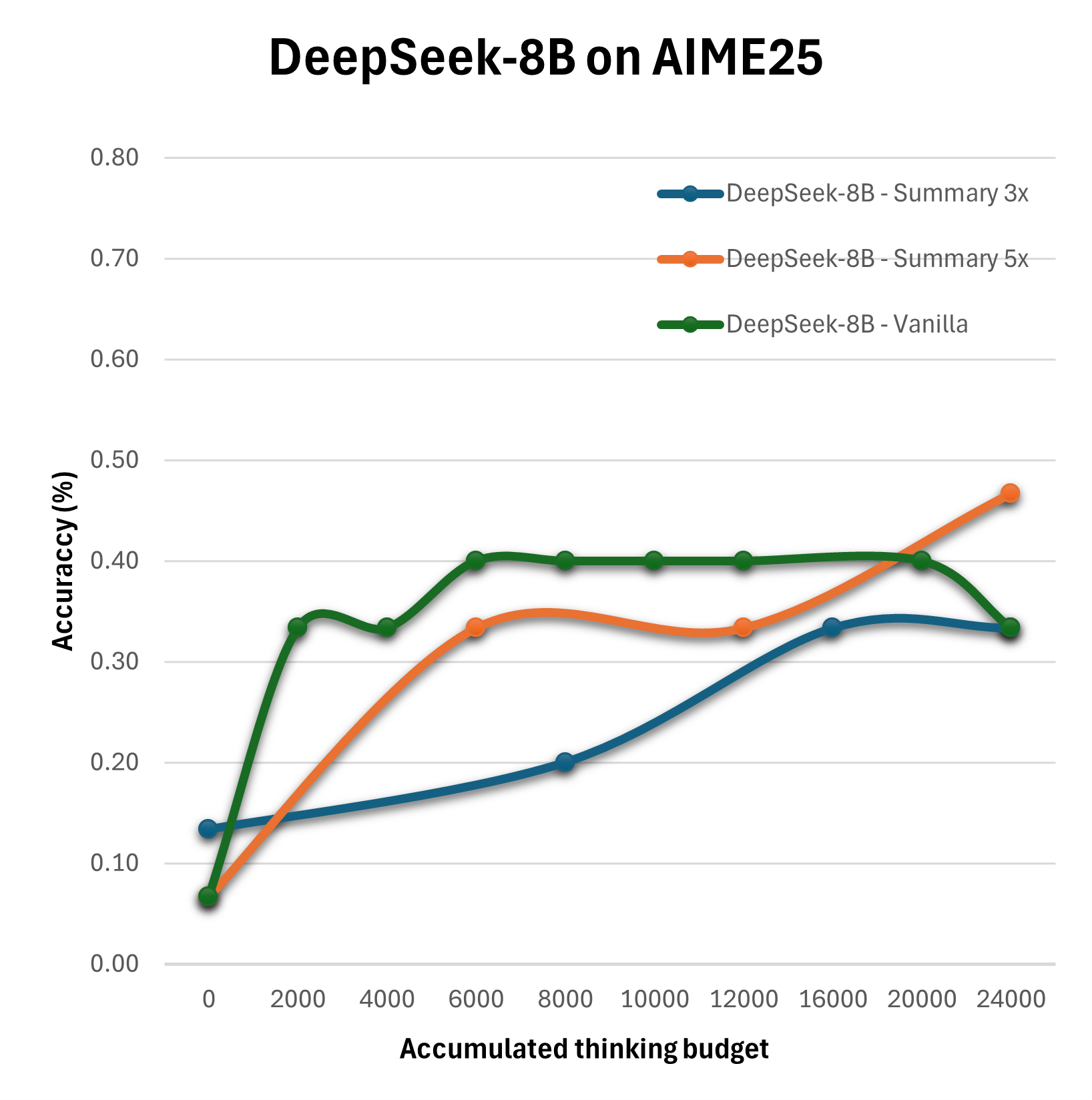}
  \end{minipage}
   \caption{Evaluation of different configurations on the AIME25 on Qwen3-8B, Qwen3-4B and DeepSeek-R1-Distill-Llama-8B.}
	\label{fig:performanceAIME25}
\end{figure*}

\end{document}